\documentclass{article}

     \PassOptionsToPackage{numbers, compress}{natbib}

\usepackage[preprint]{neurips_2019_arxiv}

\usepackage[utf8]{inputenc} 
\usepackage[T1]{fontenc}    
\usepackage[colorlinks=true, linkcolor=magenta, citecolor=magenta]{hyperref}       
\usepackage{url}            
\usepackage{booktabs}       
\usepackage{amsfonts}       
\usepackage{nicefrac}       
\usepackage{microtype}      

\usepackage{graphicx}
\usepackage{subfigure}
\usepackage{booktabs} 
\usepackage{amssymb}
\usepackage{amsmath}
\usepackage{eqparbox}
\usepackage{adjustbox}
\usepackage{multirow}
\usepackage[table,array,dvipsnames]{xcolor}
\usepackage{algorithm}
\usepackage{algorithmic}
\usepackage[labelfont={small,bf},textfont=small]{caption}
\usepackage{wrapfig,blindtext}

\usepackage{xr}
\DeclareMathOperator*{\argmax}{\arg\!\max}

\newcommand{\mn}[1]{{{#1}}}

\newcommand{\hdc}[1]{{{#1}}}

\newcommand{\beginsupplement}{%
        \setcounter{table}{0}
        \renewcommand{\thetable}{S\arabic{table}}%
        \setcounter{figure}{0}
        \renewcommand{\thefigure}{S\arabic{figure}}%
}

\def\mathunderline#1#2{\color{#1}\underline{{\color{black}#2}}\color{black}}

\title{\mn{Deep} Multi-View Learning via Task-Optimal CCA}

\author{%
  Heather D. Couture\\
  Pixel Scientia Labs, Raleigh, NC\\
  \texttt{heather@pixelscientia.com} \\
  \And
  Roland Kwitt\\
  University of Salzburg, Austria\\
  \texttt{roland.kwitt@sbg.ac.at}\\
  \AND
  J.S. Marron~~~~~ Melissa Troester~~~~~ Charles M. Perou~~~~~ Marc Niethammer\\
  University of North Carolina at Chapel Hill\\
  \texttt{marron@unc.edu}, \texttt{troester@unc.edu}, \texttt{chuck\_perou@med.unc.edu}, \texttt{mn@cs.unc.edu}\\
}

\begin{document}

\maketitle

\begin{abstract}
Canonical Correlation Analysis (CCA) is widely used for multimodal data analysis and, more recently, for discriminative tasks such as multi-view learning; however, it makes no use of class labels.  Recent CCA methods have started to address this weakness but are limited in that they do not simultaneously optimize the CCA projection for discrimination and the CCA projection itself, or they are linear only. We address these deficiencies by simultaneously optimizing a CCA-based and a task objective in an end-to-end manner. Together, these two objectives learn a non-linear CCA projection to a shared latent space that is highly correlated and discriminative. Our method shows a significant improvement over previous state-of-the-art (including deep supervised approaches) for cross-view classification, regularization with a second view, and semi-supervised learning on real data.
\end{abstract}

\section{Introduction}

\mn{CCA} is a popular data analysis technique that projects two data sources into a space in which they are maximally correlated \citep{Hotelling1936,Bie2005}. It was initially used for unsupervised data analysis to gain insights into components shared by the two sources \citep{Andrew2013,Wang2015_ICML,Wang2016}. \mn{CCA is also used to compute a} shared latent space for cross-view classification \citep{Kan2015,Wang2015_ICML,Chandar2016,Chang2018}, for representation learning on multiple views that are then joined for prediction \citep{Sargin2007,Dorfer2016}, and for classification from a single view when a second view is available during training \citep{Arora2012}. While some of the correlated \mn{CCA} features are useful for discriminative tasks, many represent properties that are of no use for classification and obscure correlated information that is beneficial. This problem is magnified with recent non-linear extensions of CCA, implemented via neural networks \mn{(NNs)}, that make significant strides in improving correlation \citep{Andrew2013,Wang2015_ICML,Wang2016,Chang2018} but often at the expense of discriminative capability (cf. \S\ref{sec_mnist}).  Therefore, we present a new deep learning technique to project the data from two views to a shared space that is also discriminative.

Most prior work that boosts the discriminative capability of CCA is \emph{linear only} \citep{Lee2015,Singanamalli2014,Duan2016}. More recent work using NNs still remains limited in that it optimizes discriminative capability for an intermediate representation rather than the final CCA projection \citep{Dorfer2016}, or optimizes the CCA objective only during pre-training, not while training the task objective \citep{Dorfer2018}. We advocate to jointly optimize CCA and a discriminative objective by computing the CCA projection within a network layer \mn{while} applying a task-driven operation such as classification. Experimental \mn{results show} that our method significantly improves upon previous work \citep{Dorfer2016,Dorfer2018} due to its focus on both the shared latent space and a task-driven objective. The latter is particularly important on small training set sizes.  

While alternative approaches to multi-view learning via CCA exist, they \mn{typically focus} on a reconstruction objective. That is, they transform the input into a shared space such that the input could be reconstructed -- either individually, or reconstructing one view from the other. Variations of coupled dictionary learning \citep{Shekhar2014,Xu2015,Cha2015,Bahrampour2015} and autoencoders \citep{Wang2015_ICML,Bhatt2017} have been used in this context. CCA-based objectives, such as the model used in this work, instead learn a transformation to a shared space without the need for reconstructing the input. This task may be easier and sufficient in producing a representation for multi-view classification \citep{Wang2015_ICML}. We show that the CCA objective can equivalently be expressed as an $\ell_2$ distance minimization in the shared space plus an orthogonality constraint. {Orthogonality constraints help regularize NNs} \citep{Huang2018_dbn}; we present three techniques to accomplish this. While our method is derived from CCA, \mn{by} manipulating the orthogonality constraints, \mn{we obtain deep CCA approaches that compute} a shared latent space that is also discriminative.

Overall, our method enables end-to-end training via mini-batches, and \mn{we demonstrate the effectiveness of our model for three different tasks:} \mn{1)} \hdc{cross-view classification on a} variation of MNIST \citep{Lecun1998} showing significant improvements in accuracy, \mn{2) \hdc{regularization when two views are available for training but only one at test time on} a cancer imaging and genomic data set with only 1,000 samples, and 3) semi-supervised representation learning to improve speech recognition.} In addition, our approach is more robust in \mn{the} small sample size regime than alternative methods. \mn{Our} experiments on real data show the effectiveness of our method in learning a shared space that is more discriminative than \mn{current} state-of-the-art methods \mn{for a variety of tasks}.


\section{Background}
\label{sec_background}

\mn{We first introduce CCA and present our task-driven approach in \S\ref{sec_methods}.} \mn{Linear and non-linear CCA are unsupervised} and find the shared signal between a pair of data sources, by maximizing the sum correlation between corresponding projections. Let $\mathbf{X}_1 \in \mathbb{R}^{d_1 \times n}$ and $\mathbf{X}_2 \in \mathbb{R}^{d_2 \times n}$ be mean-centered input data from two different views with $n$ samples and $d_1$, $d_2$ features, respectively.

\noindent
\textbf{CCA.} The objective is to maximize the correlation between 
$\mathbf{a}_1 = \mathbf{w}_1^\top \mathbf{X}_1$ and $\mathbf{a}_2 = \mathbf{w}_2^\top \mathbf{X}_2$, where $\mathbf{w}_1$ and 
$\mathbf{w}_2$ are projection vectors \citep{Hotelling1936}. The first canonical directions are found via
\begin{equation*}
  \argmax_{\mathbf{w}_1,\mathbf{w}_2} \text{corr}\big(\mathbf{w}_1^\top \mathbf{X}_1, \mathbf{w}_2^\top \mathbf{X}_2\big)
  \label{eqn_cca}
\end{equation*}
and subsequent projections are found by maximizing the same correlation but in orthogonal directions. 
Combining the projection vectors into matrices $\mathbf{W}_1 = [\mathbf{w}_1^{(1)},\ldots,\mathbf{w}_1^{(k)}]$ and $\mathbf{W}_2 = [\mathbf{w}_2^{(1)},\ldots,\mathbf{w}_2^{(k)}]$ ($k \leq \text{min}(d_1,d_2)$), CCA can be reformulated as a trace maximization under orthonormality constraints on the projections, i.e.,
\begin{equation}
  \argmax_{\mathbf{W}_1,\mathbf{W}_2} \text{tr}( \mathbf{W}_1^\top \boldsymbol{\Sigma}_{12} \mathbf{W}_2 ) ~~~~ \text{s.t. } \mathbf{W}_1^\top \boldsymbol{\Sigma}_{1} \mathbf{W}_1 = \mathbf{W}_2^\top \boldsymbol{\Sigma}_{2} \mathbf{W}_2 = \mathbf{I}
  \label{eqn_cca_trace}
\end{equation}
for covariance matrices $\boldsymbol{\Sigma}_1=X_1 X_1^T$, $\boldsymbol{\Sigma}_2=X_2 X_2^T$, and cross-covariance matrix 
$\boldsymbol{\Sigma}_{12}=X_1 X_2^T$.
Let $\mathbf{T} = \boldsymbol{\Sigma}_{1}^{-1/2} \boldsymbol{\Sigma}_{12} \boldsymbol{\Sigma}_{2}^{-1/2}$ and its singular value decomposition (SVD) be $\mathbf{T} = \mathbf{U}_1 \text{diag}(\boldsymbol{\sigma}) \mathbf{U}_2^\top$ with singular values $\boldsymbol{\sigma}=[\sigma_1,\ldots,\sigma_{\text{min}(d_1,d_2)}]$ in descending order. $\mathbf{W}_1$ and $\mathbf{W}_2$ are computed from the top $k$ singular vectors of $\mathbf{T}$ as
$\mathbf{W}_1 = \boldsymbol{\Sigma}_{1}^{-1/2} \mathbf{U}_1^{(1:k)}$ and $\mathbf{W}_2 = \boldsymbol{\Sigma}_{2}^{-1/2} \mathbf{U}_2^{(1:k)}$ 
where $\mathbf{U}^{(1:k)}$ denotes the $k$ first columns of matrix $\mathbf{U}$. The sum correlation in the projection space is equivalent to 
\begin{equation}
  \sum_{i=1}^k \text{corr}\big(\big(\mathbf{w}_1^{(i)}\big)^\top X_1, \big(\mathbf{w}_2^{(i)})^\top \mathbf{X}_2 \big) = \sum_{i=1}^{k} \sigma_i^2\enspace,
  \label{eqn_cca_corr}
\end{equation}
i.e., the sum of the top $k$ singular values. A regularized variation of CCA (RCCA) ensures that the covariance matrices are positive definite by computing the covariance matrices as $\hat{\boldsymbol{\Sigma}}_{1} = \frac{1}{n-1} \mathbf{X}_1 \mathbf{X}_1^\top + r\mathbf{I}$ and $\hat{\boldsymbol{\Sigma}}_{2} = \frac{1}{n-1} \mathbf{X}_2 \mathbf{X}_2^\top + r\mathbf{I}$, for regularization parameter $r > 0$ and identity matrix $\mathbf{I}$ \citep{Bilenko2016}.

\begin{figure}[t!]
\centering
\includegraphics[width=0.99\textwidth]{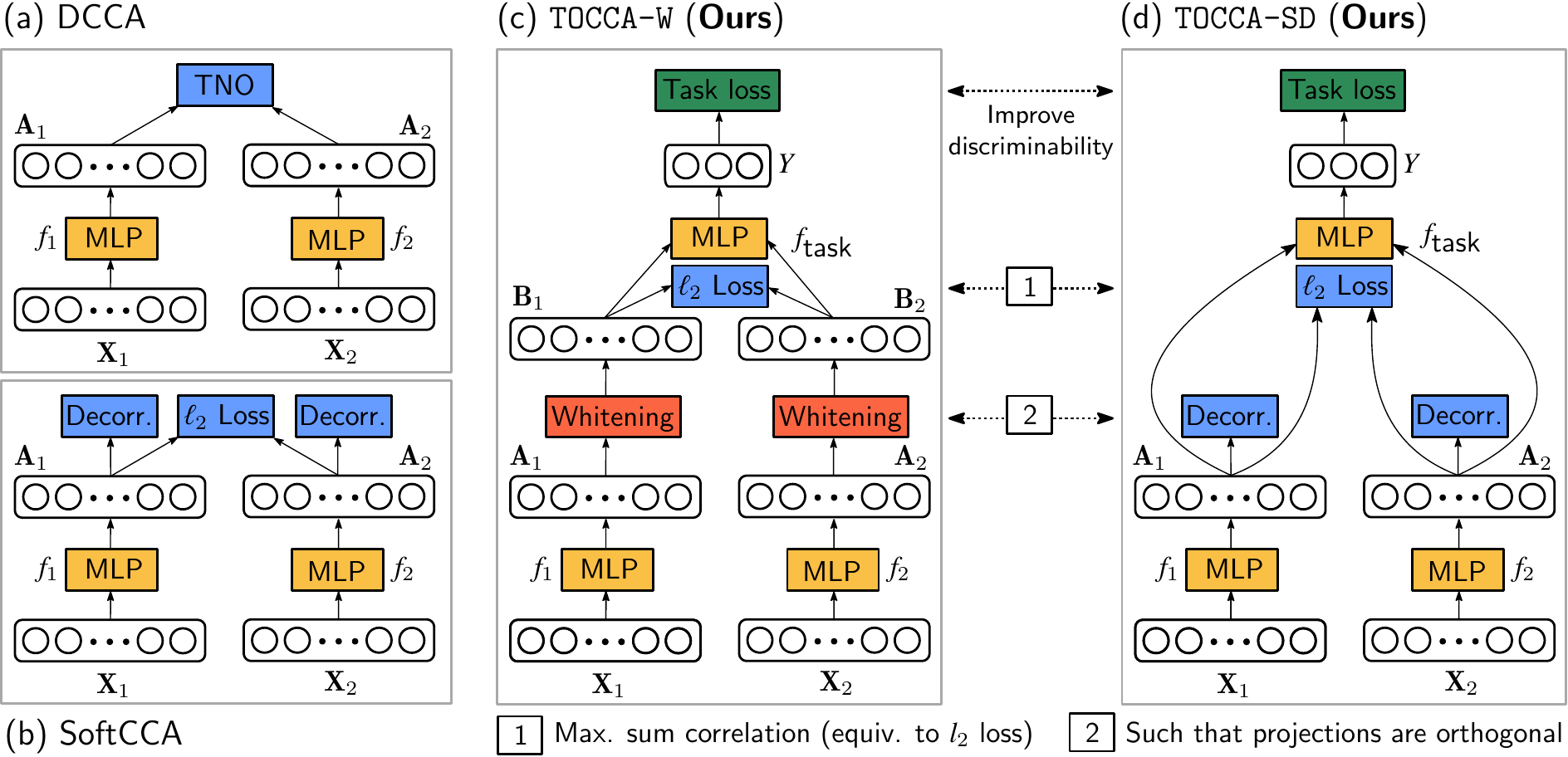}
\caption{Deep CCA architectures: (a) DCCA maximizes the sum correlation in projection space \mn{by optimizing} an equivalent loss, the trace norm objective (TNO) \citep{Andrew2013}; (b) SoftCCA relaxes the orthogonality constraints by regularizing with \mn{soft decorrelation (Decorr)} and optimizes the $\ell_2$ distance in the projection space (\mn{equivalent to sum correlation with activations normalized to unit variance}) \citep{Chang2018}. Our \texttt{TOCCA} methods add a task loss and apply CCA orthogonality constraints by regularizing in two ways: (c) \texttt{TOCCA-W} uses whitening and (d) \texttt{TOCCA-SD} uses Decorr.  The third method that we propose, \texttt{TOCCA-ND}, simply removes the Decorr components of \texttt{TOCCA-SD}.\vspace{-0.4cm}}
  \label{fig_network_diagram}
\end{figure}

\noindent
\textbf{DCCA}. Deep CCA adds non-linear projections to CCA by non-linearly mapping the input via a multilayer perceptron (MLP). In particular, inputs $\mathbf{X}_1$ and $\mathbf{X}_2$ are mapped via non-linear functions $f_1$ and $f_2$, parameterized by $\theta_1$ and $\theta_2$, resulting in activations $\mathbf{A}_1 = f_1(\mathbf{X}_1;\theta_1)$ and $\mathbf{A}_2 = f_2(\mathbf{X}_2;\theta_2)$ (assumed to be mean centered) \citep{Andrew2013}.
When implemented by a NN, $\mathbf{A}_1$ and $\mathbf{A}$ are the output activations of the final 
layer with $d_o$ features. Fig.~\ref{fig_network_diagram}(a) shows the network structure.
DCCA optimizes the same objective as CCA, see Eq.~\eqref{eqn_cca_trace}, but using activations $\mathbf{A}_1$ and $\mathbf{A}_2$. Regularized covariance matrices are computed accordingly and 
the solution for $\mathbf{W}_1$ and $\mathbf{W}_2$ can be computed using SVD just as with linear CCA.
When $k = d_o$ (i.e., the number of CCA components is equal to the number of features in $\mathbf{A}_1$ and $\mathbf{A}_2$), optimizing the sum correlation in the projection space, as in Eq.~\eqref{eqn_cca_corr}, is equivalent to optimizing the following matrix \emph{trace norm objective (TNO)}
\begin{equation*}
  \mathcal{L}_{\text{TNO}}(\mathbf{A}_1,\mathbf{A}_2) = \|\mathbf{T}\|_{\text{tr}} = \text{tr}\big(\mathbf{T}^\top \mathbf{T}\big)^{1/2}\enspace,
\end{equation*}
where $\mathbf{T} = \boldsymbol{\Sigma}_{1}^{-1/2} \boldsymbol{\Sigma}_{12} \boldsymbol{\Sigma}_{2}^{-1/2}$
as in case of CCA \citep{Andrew2013}.  DCCA optimizes this objective directly, \emph{without} a need to compute the CCA projection within the network. The TNO is optimized first, followed by a linear CCA operation before downstream tasks like classification are performed.

\noindent
\textbf{SoftCCA.} While DCCA enforces orthogonality constraints on projections $\mathbf{W}_1^\top \mathbf{A}_1$ and $\mathbf{W}_2^\top \mathbf{A}_2$, SoftCCA relaxes them using regularization \citep{Chang2018}.  Final projection matrices $\mathbf{W}_1$ and $\mathbf{W}_2$ are integrated into $f_1$ and $f_2$ as the top network layer. The trace objective for DCCA in Eq.~\eqref{eqn_cca_trace} can be rewritten as minimizing the $\ell_2$ distance between the projections when each feature in $\mathbf{A}_1$ and $\mathbf{A}_2$ is normalized to a unit variance \citep{Li2003}, leading to\footnote{We use this $\ell_2$ distance objective in our formulation.
}
  $\mathcal{L}_{\ell_2 \text{ dist}}(A_1,A_2) = \|\mathbf{A}_1 - \mathbf{A}_2 \|_F^2\enspace.$
Regularization in SoftCCA penalizes the off-diagonal elements of the covariance matrix $\boldsymbol{\Sigma}$, \hdc{using a running average computed over batches as $\hat{\boldsymbol{\Sigma}}$ and a loss of}
  $\mathcal{L}_{\text{Decorr}}(\mathbf{A}) = \sum_{i \neq i}^{d_o} |\hat{\Sigma}_{i,j}|$.
Overall, the SoftCCA loss takes the form
\begin{equation*}
  \mathcal{L}_{\ell_2 \text{ dist} }(\mathbf{A}_1,\mathbf{A}_2) + \lambda \big( \mathcal{L}_{\text{Decorr}}(\mathbf{A}_1) + \mathcal{L}_{\text{Decorr}}(\mathbf{A}_2) \big)\enspace.
\end{equation*}

\noindent
\textbf{Supervised CCA methods.} CCA, DCCA, and SoftCCA are all unsupervised methods to learn a projection to a shared space in which the data is maximally correlated.  Although these methods have shown utility for discriminative tasks, a CCA decomposition \mn{may not be} optimal for classification because \mn{features that are correlated may not be} discriminative. Our experiments will \mn{show} that maximizing the correlation objective too much can degrade performance on discriminative tasks.

CCA has previously been extended to supervised settings by maximizing the total correlation between each view and the training labels in addition to each pair of views \citep{Lee2015,Singanamalli2014}, and by maximizing the separation of classes \citep{Kan2015,Dorfer2016}. Although these methods incorporate the class labels, they do not directly optimize for classification. Dorfer et. al's CCA Layer (CCAL) is the closest to our method.  It optimizes a task loss operating on a CCA projection; however, the CCA objective itself is only optimized during pre-training, not in an end-to-end manner \citep{Dorfer2018}. Other supervised CCA methods are linear only \citep{Singanamalli2014,Lee2015,Kan2015,Duan2016}. Instead of computing the CCA projection within the network, as in CCAL, we optimize the non-linear mapping into the shared space \emph{together} with the CCA part.

\section{Task-Optimal CCA (TOCCA)}
\label{sec_methods}

\mn{To compute} a shared latent space that is also discriminative, we start with the DCCA formulation and add a task-driven term to the optimization objective. The CCA component finds features that are correlated between views, while the task component ensures that they are also discriminative. This model can be used for representation learning on multiple views before joining representations for prediction \citep{Sargin2007,Dorfer2016} and for classification when two views are available for training but only one at test time \citep{Arora2012}. In \S\ref{sec_experiments}, we demonstrate both use cases on real data. Our methods and related \mn{NN} models from the literature are summarized in \mn{Tab.~}\ref{tbl_methods} (suppl. material); \mn{Fig.~\ref{fig_network_diagram} shows schematic diagrams.}

While DCCA \mn{optimizes} the sum correlation through an equivalent loss function (TNO), the CCA projection itself is computed only {\it after} optimization. Hence, the projections cannot be used to optimize another task simultaneously. The main challenge in developing a task-optimal form of deep CCA that discriminates based on the CCA projection is in computing this projection within the network -- a necessary step to enable simultaneous training of both objectives.  We tackle this by focusing on the two components of DCCA: maximizing the sum correlation between activations $\mathbf{A}_1$ and $\mathbf{A}_2$ and enforcing orthonormality constraints within $\mathbf{A}_1$ and $\mathbf{A}_2$. We achieve both by transforming the CCA objective and present three methods that progressively relax the orthogonality constraints.

We further improve upon DCCA by enabling \mn{mini-batch computations} for improved flexibility and test 
performance. DCCA was developed \mn{for large} batches because correlation is not separable across batches. While large batch implementations of stochastic gradient optimization \mn{can increase computational efficiency via parallelism}, small batch training provides more up-to-date gradient calculations, \mn{allowing a wider range of learning} rates and improving test accuracy \citep{Masters2018}. \mn{We reformulate} the correlation objective as the $\ell_2$ distance (following SoftCCA), \mn{enabling separability across batches}. We ensure a normalization to one \mn{via} batch normalization without the scale and shift parameters \citep{Ioffe2015}.

\textbf{Task-driven objective.} First, we apply non-linear functions $f_1$ and $f_2$ (via MLPs) 
to each view $\mathbf{X}_1$ and $\mathbf{X}_2$, i.e., $\mathbf{A}_1 = f_1(\mathbf{X}_1;\theta_1)$ and 
$\mathbf{A}_2 = f_2(\mathbf{X}_2;\theta_2)$. Second, a task-specific function $f_{\text{task}}(\mathbf{A};\theta_{\text{task}})$ operates on the outputs $\mathbf{A}_1$ and $\mathbf{A}_2$. In particular, $f_1$ and $f_2$ are optimized so that the $\ell_2$ distance between $\mathbf{A}_1$ and $\mathbf{A}_2$ is minimized; therefore, $f_{\text{task}}$ can be trained to operate on both inputs $\mathbf{A}_1$ and $\mathbf{A}_2$. We combine CCA and task-driven objectives as a weighted sum with a hyperparameter for tuning. This model is flexible, in that the task-driven goal can be used for classification \citep{Krizhevsky2012,Dorfer2016_lda}, regression \citep{Katzman2016}, clustering \citep{Caron2018}, or any other task. \mn{See Tab.~\ref{tbl_methods} (suppl. material) for an overview.}

\noindent
\textbf{Orthogonality constraints.}   The remaining complications for mini-batch optimization are the orthogonality constraints, for which we propose three solutions, each handling the orthogonality constraints of CCA in a different way\mn{: whitening, soft decorrelation, and no decorrelation.}  

\noindent
\textbf{1) Whitening (\texttt{TOCCA-W}).} CCA applies orthogonality constraints to $\mathbf{A}_1$ and $\mathbf{A}_2$.  We accomplish this with a linear \mn{whitening} transformation that transforms the activations such that their covariance becomes the identity matrix, \mn{i.e.}, features are uncorrelated. Decorrelated Batch Normalization (DBN) has previously been used to regularize deep models by decorrelating features \citep{Huang2018_dbn} and inspired our solution. In particular, we apply a transformation 
$\mathbf{B} = \mathbf{U}\mathbf{A}$ to make $\mathbf{B}$ orthonormal, i.e., $\mathbf{B}\mathbf{B}^\top = 
\mathbf{I}$.

We use a Zero-phase Component Analysis (ZCA) whitening \mn{transform composed} of three steps: rotate the data to decorrelate it, rescale each axis, and rotate back to the original space.  Each of these transformations is learned from the data. 
Any matrix $\mathbf{U} \epsilon \mathbb{R}^{d_o \times d_o}$ \mn{satisfying} $\mathbf{U}^\top 
\mathbf{U} = \boldsymbol{\Sigma}^{-1}$ whitens the data, where $\boldsymbol{\Sigma}$ denotes the covariance
matrix of $\mathbf{A}$. \mn{As} $\mathbf{U}$ is only defined up to a rotation, \mn{it} is not unique. PCA whitening follows the first two steps and uses the eigendecomposition of $\boldsymbol{\Sigma}$: $\mathbf{U}_{PCA} = \boldsymbol{\Lambda}^{-1/2} \mathbf{V}^\top$ for $\boldsymbol{\Lambda}=\text{diag}(\lambda_1,\ldots,\lambda_{d_o})$ and $\mathbf{V}=[\mathbf{v}_1,\ldots,\mn{\mathbf{v}_{d_o}}]$, where $(\lambda_i,\mathbf{v}_i)$ are the eigenvalue, eigenvector pairs of $\boldsymbol{\Sigma}$. \mn{As} PCA whitening suffers from stochastic axis swapping, \mn{neurons} are not stable \mn{between batches} \citep{Huang2018_dbn}.  
ZCA whitening uses the transformation $\mathbf{U}_{\text{ZCA}} = \mathbf{V} \boldsymbol{\Lambda}^{-1/2} \mathbf{V}^T$ in which PCA whitening is first applied, followed by a rotation back to the original space.  Adding the rotation $\mathbf{V}$ brings the whitened data $\mathbf{B}$ as close as possible to the original data $\mathbf{A}$ \citep{Kessy2015}.

Computation of $\mathbf{U}_{\text{ZCA}}$ is \mn{clearly depend on} $\boldsymbol{\Sigma}$. 
While Huang et al. \citep{Huang2018_dbn} used a running average \mn{of $\mathbf{U}_{\text{ZCA}}$} over batches, we apply this stochastic approximation to 
$\boldsymbol{\Sigma}$ for each view using the update
 $ \boldsymbol{\Sigma}^{(k)} = \alpha \boldsymbol{\Sigma}^{(k-1)} + (1-\alpha) \boldsymbol{\Sigma}^b$
for batch $k$ where $\boldsymbol{\Sigma}^b$ is the covariance matrix for the current batch and $\alpha \in(0,1)$ is the momentum.  We then compute the ZCA transformation from $\boldsymbol{\Sigma}^{(k)}$ to do whitening as
  $\mathbf{B} = f_{\text{ZCA}}(\mathbf{A}) = \mathbf{U}_{\text{ZCA}}^{(k)} \mathbf{A}$.
At test time, $\mathbf{U}^{(k)}$ from the last training batch is used. Algorithm \ref{algo_zca} 
(suppl. material) describes ZCA whitening in greater detail.  
In summary, \texttt{TOCCA-W} integrates both the correlation and task-driven objectives, with decorrelation performed by whitening, into
\begin{equation*}
  \mathcal{L}_{\text{task}}(f_{\text{task}}(\mathbf{B}_1),Y) + \mathcal{L}_{\text{task}}(f_{\text{task}}(\mathbf{B}_2),Y) + \lambda~ \mathcal{L}_{\ell_2 \text{ dist} }(\mathbf{B}_1,\mathbf{B}_2)\enspace,
\end{equation*}
where $\mathbf{B}_1$ and $\mathbf{B}_2$ are whitened outputs of $\mathbf{A}_1$ and $\mathbf{A}_2$, respectively.

\noindent
\textbf{2) Soft decorrelation (\texttt{TOCCA-SD}).}  While fully independent components may be beneficial in regularizing NNs on some data sets, a softer decorrelation may be more suitable on others. In this second formulation we relax the orthogonality constraints using regularization, following the Decorr loss of SoftCCA \citep{Chang2018}. The loss function for this formulation is
\begin{multline*}
  \mathcal{L}_{\text{task}}(f_{\text{task}}(\mathbf{A}_1),Y) + \mathcal{L}_{\text{task}}(f_{\text{task}}(\mathbf{A}_2),Y) + \lambda_1 \mathcal{L}_{\ell_2 \text{ dist} }(\mathbf{A}_1,\mathbf{A}_2) + \lambda_2 \big( \mathcal{L}_{\text{Decorr}}(\mathbf{A}_1) + \mathcal{L}_{\text{Decorr}}(\mathbf{A}_2) \big)\enspace.
\end{multline*}

\noindent
\textbf{3) No decorrelation (\texttt{TOCCA-ND}).}  When CCA is used in an unsupervised manner, some form of orthogonality constraint or decorrelation is necessary to ensure that $f_1$ and $f_2$ do not simply produce multiple copies of the same feature.  While this result could maximize the sum correlation, it is not helpful in capturing useful projections. In the task-driven setting, the discriminative term ensures that the features in $f_1$ and $f_2$ are not replicates of the same information. \texttt{TOCCA-ND} \mn{therefore} removes the decorrelation term entirely, forming the simpler objective
\begin{equation*}
  \mathcal{L}_{\text{task}}(f_{\text{task}}(\mathbf{A}_1),Y) + \mathcal{L}_{\text{task}}(f_{\text{task}}(\mathbf{A}_2),Y) + \lambda \mathcal{L}_{\ell_2 \text{ dist} }(\mathbf{A}_1,\mathbf{A}_2)\enspace.
\end{equation*}
These three models allow \mn{testing} whether \mn{whitening or soft decorrelation benefit} a task-driven model.

\noindent
\textbf{Computational complexity.} Due to the eigendecomposition, \texttt{TOCCA-W} has a complexity of $O(d_o^3)$ compared to $O(d_o^2)$ for \texttt{TOCCA-SD}, with respect to output dimension $d_o$.  However, $d_o$ is typically small ($\leq 100$) and this extra computation is only performed once per batch.  The difference in runtime is less than 6.5\% for a batch size of 100 or 9.4\% for a batch size of 30 (Table \ref{tbl_runtime}, suppl. material).

In summary, all three variants are motivated by adding a task-driven component to deep CCA. 
\texttt{TOCCA-ND} is the most relaxed \mn{and directly} attempts to obtain identical latent representations. Experiments will show that whitening (\texttt{TOCCA-W}) and soft decorrelation (\texttt{TOCCA-SD}) provide a beneficial regularization. Further, since the $\ell_2$ distance that we optimize was shown to be equivalent to the sum correlation \mn{(cf. \S\ref{sec_background} SoftCCA paragraph)}, all three \texttt{TOCCA} models maintain the goals of CCA just with different relaxations of the orthogonality constraints. \mn{See Tab.~\ref{tbl_methods} (suppl. material) for an overview.}

\section{Experiments}
\label{sec_experiments}

We validated our methods on three different data sets: MNIST handwritten digits, the Carolina Breast Cancer Study (CBCS) using imaging and genomic features, and speech data from the Wisconsin X-ray Microbeam Database (XRMB).  Our experiments show the utility of our methods for (1) cross-view classification, (2) regularization with a second view during training when only one view is available at test time, and (3) representation learning on multiple views that are joined for prediction.

\noindent
\textbf{Implementation.\footnote{{\footnotesize Code will be available on GitHub soon.}}} Each layer of our network consists of a fully connected layer, followed by a ReLU activation and batch normalization \citep{Ioffe2015}. We used the Nadam optimizer and tuned hyperparameters on a validation set via random search; settings and ranges are specified in Table \ref{tbl_hyperparameters} (suppl. material).  We used Keras with the Theano backend and an Nvidia GeForce GTX 1080 Ti. Our implementations of DCCA, SoftCCA, and Joint DCCA/DeepLDA \citep{Dorfer2016} also use ReLu activation and batch normalization. We modified CCAL-$\mathcal{L}_{\text{rank}}$ \citep{Dorfer2018} to use a softmax function and cross-entropy loss for classification, instead of a pairwise ranking loss for retrieval, referring to this modification as CCAL-$\mathcal{L}_{\text{ce}}$. 

\subsection{Cross-view classification on MNIST digits}
\label{sec_mnist}

We formed a multi-view data set from the MNIST handwritten digit image data set \citep{Lecun1998}. Following Andrew et al. \citep{Andrew2013}, we split each $28 \times 28$ image in half horizontally, creating left and right views that are each $14 \times 28$ pixels. All images were flattened into a vector with 392 features. 
The full data set consists of 60k training images and 10k test images. We used a random set of up to 50k for training and the remaining training images for validation. We used the full 10k image test set.

We evaluated cross-view classification accuracy by first computing the projection for each view, then we trained a linear SVM on one view's projection, and finally we used the other view's projection at test time. While the task-driven methods presented in this work learn a classifier within the model, this test setup enables a fair comparison with the unsupervised CCA variants and validates the discriminativity of the features learned. Notably, using the built-in softmax classifier performed similarly to the SVM (not shown), as much of the power of our methods comes from the representation learning part. \hdc{We do not compare with a simple supervised NN because this setup does not learn the shared space necessary for cross-view classification.}  We report \hdc{results averaged} over five randomly selected training/validation sets; the test set always remained the same.

\begin{figure}
\centering
\setlength{\tabcolsep}{0pt}
\includegraphics[height=2.15in]{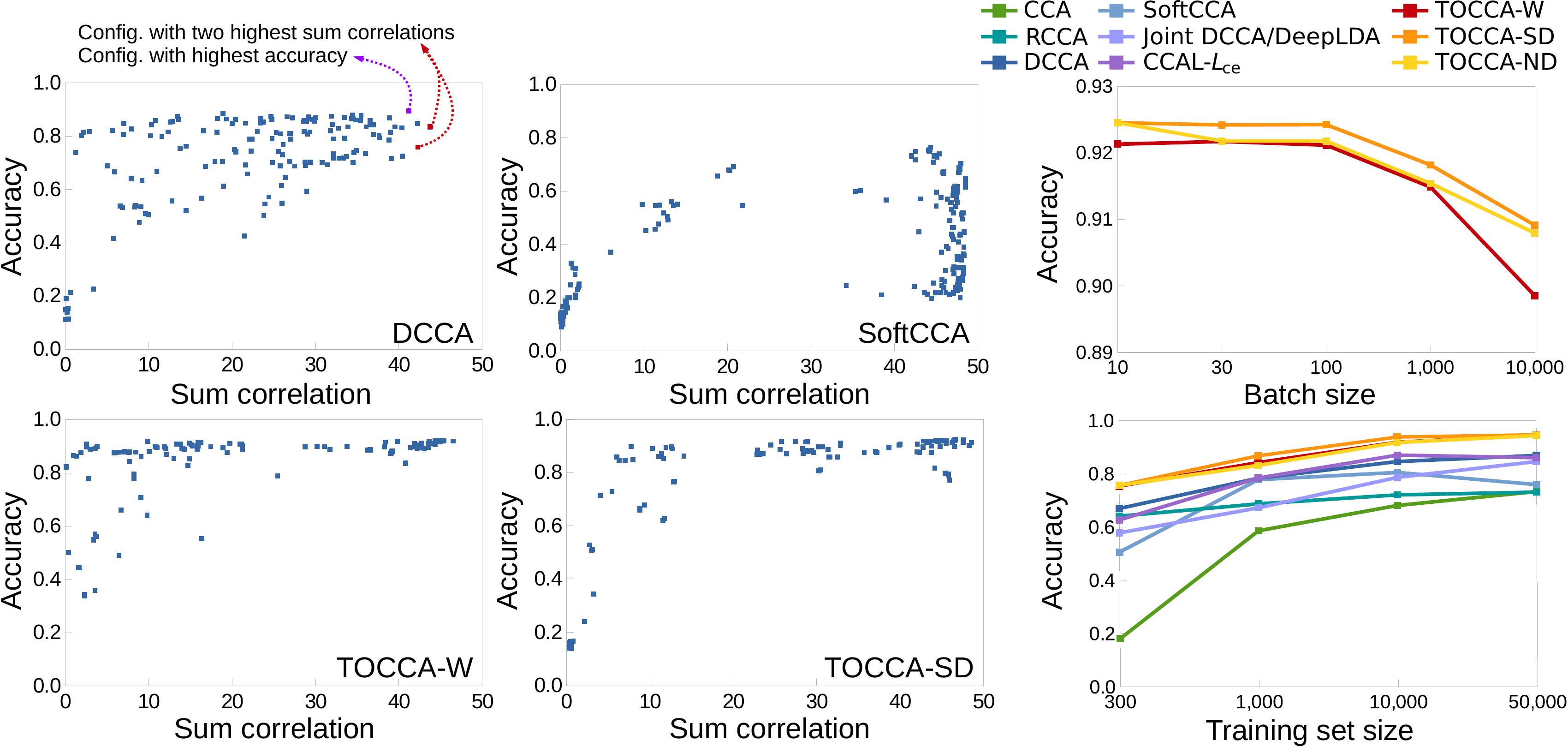}
\caption{\textit{Left}: Sum correlation vs. cross-view classification accuracy (on MNIST) across different hyperparameter settings on a training set size of 10,000 for DCCA \citep{Andrew2013}, SoftCCA \citep{Chang2018}, \hdc{\texttt{TOCCA-W}, and \texttt{TOCCA-SD}}. For unsupervised methods \hdc{(DCCA and SoftCCA)}, large correlations do not necessarily imply good accuracy. \textit{Right}: The effect of batch size on classification accuracy for each \texttt{TOCCA} method on MNIST (training set size of 10,000), and the effect of training set size on classification accuracy for each method. Our \texttt{TOCCA} variants out-performed all others across all training set sizes.
\label{fig_mnist_overall}\vspace{-0.75cm}}
\end{figure}

\noindent
\textbf{Correlation vs. classification accuracy}  We first demonstrate the importance of adding a task-driven component to DCCA by showing that maximizing the sum correlation between views is not sufficient. Fig.~\ref{fig_mnist_overall} (\emph{left}) shows the sum correlation vs. cross-view classification accuracy across many different hyperparameter settings for DCCA \citep{Andrew2013}, SoftCCA \citep{Chang2018}, \hdc{and \texttt{TOCCA}}. We used 50 components for each; thus, the maximum sum correlation was 50. The sum correlation was measured after applying linear CCA to ensure that components were independent. With DCCA a larger correlation tended to produce a larger classification accuracy, but there was still a large variance in classification accuracy amongst hyperparameter settings that produced a similar sum correlation.  For example, with the two farthest right points in the plot (colored \textcolor{red}{red}), their classification accuracy differs by 10\%, and they are not even the points with the best classification accuracy (colored \textcolor{magenta}{purple}).
The pattern is different for SoftCCA. There was an increase in classification accuracy as sum correlation increased but only up to a point. For higher sum correlations, the classification accuracy varied even more from 20\% to 80\%.  Further experiments (not shown) have indicated that when the sole objective is correlation, some of the projection directions are simply not discriminative, particularly when there are a large number of classes. Hence, optimizing for sum correlation alone does not guarantee a discriminative model. \hdc{\texttt{TOCCA-W} and \texttt{TOCCA-SD} show a much greater classification accuracy across a wide range of correlations and, overall, the best accuracy when correlation is greatest.}

\noindent
\textbf{Effect of batch size.}  
Fig.~\ref{fig_mnist_overall} (\emph{right}) plots the batch size vs. classification accuracy for a training set size of $10,000$.  \hdc{We tested batch sizes from $10$ to $10{,}000$; a batch size of 10 or 30 was best} for all three variations of \texttt{TOCCA}. This is in line with previous work that found the best performance with a batch size between 2 and 32 \citep{Masters2018}.  We used a batch size of 32 in the remaining experiments on MNIST.

\noindent
\textbf{Effect of training set size.}  We manipulated the training set size in order to study the robustness of our methods. In particular, Fig.~\ref{fig_mnist_overall} (\emph{right}) shows the cross-view classification accuracy for training set sizes from $n=300$ to $50{,}000$.  While we expected that performance would decrease for smaller training set sizes, some methods were more susceptible to this degradation than others. The classification accuracy with CCA dropped significantly for $n=300$ and $1{,}000$, due to overfitting and instability issues related to the covariance and cross-covariance matrices. SoftCCA shows similar behavior (prior work \citep{Chang2018} on this method did not test such small training set sizes).

\begin{figure}
  \centering
  \setlength\tabcolsep{0.01in}
  \begin{tabular}{ccccc|ccc}
    \small CCA & \small RCCA & \small DCCA & \small SoftCCA & \small CCAL-$\mathcal{L}_{\text{ce}}$ & \small \texttt{TOCCA-W} &
    \multirow{2}{*}{\includegraphics[height=0.92in,trim={390px 80px 0 80px},clip]{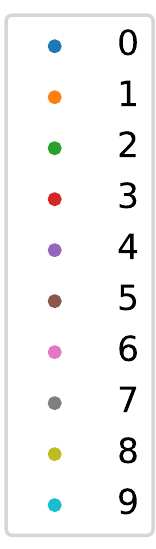}} \\
    \includegraphics[height=0.69in,trim={0 0 45px 0},clip]{figures/sne_sub10000_CCA_digits-crop.pdf} &
    \includegraphics[height=0.69in,trim={0 0 45px 0},clip]{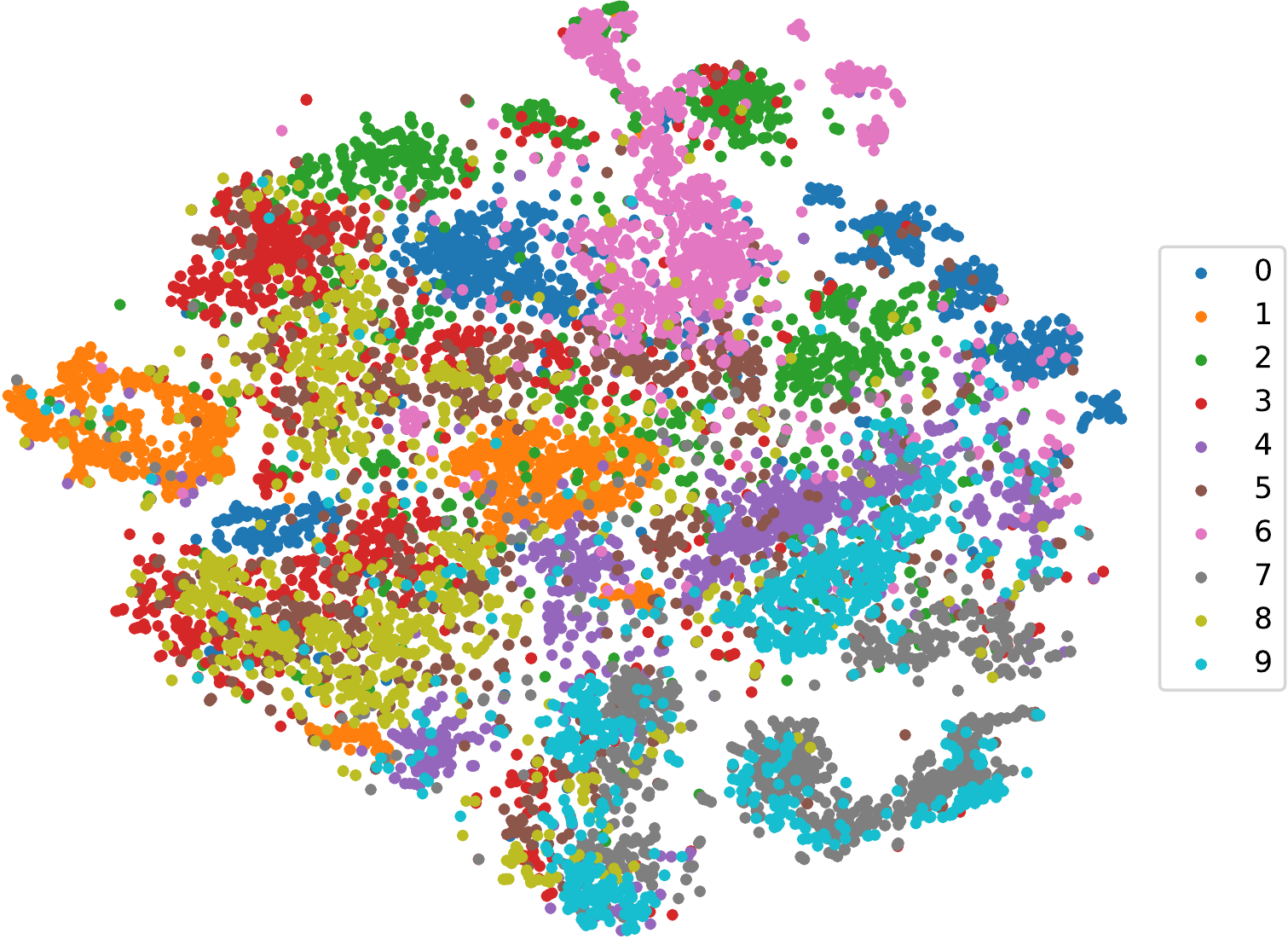} &
    \includegraphics[height=0.69in,trim={0 0 45px 0},clip]{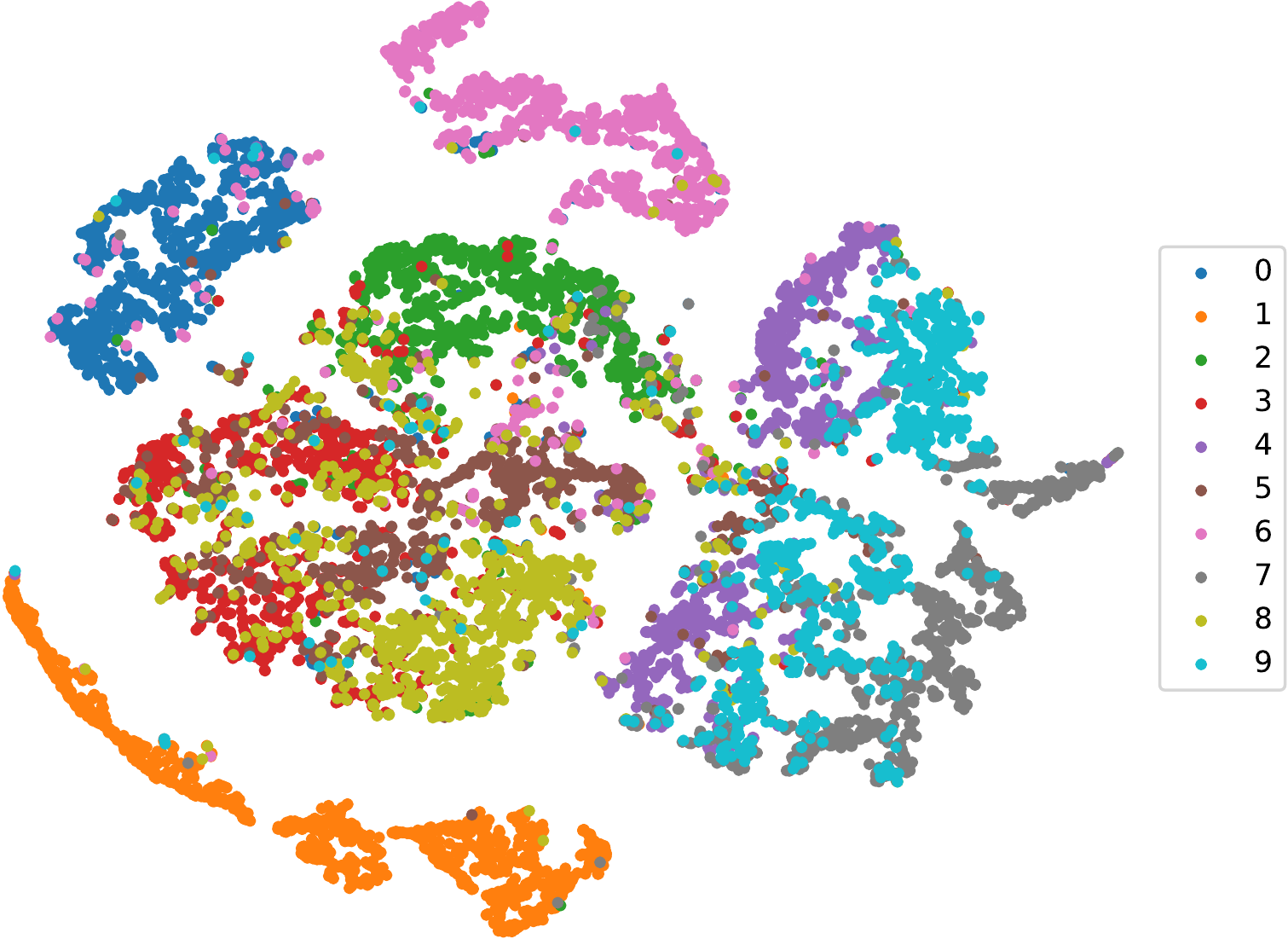} &
    \includegraphics[height=0.69in,trim={0 0 45px 0},clip]{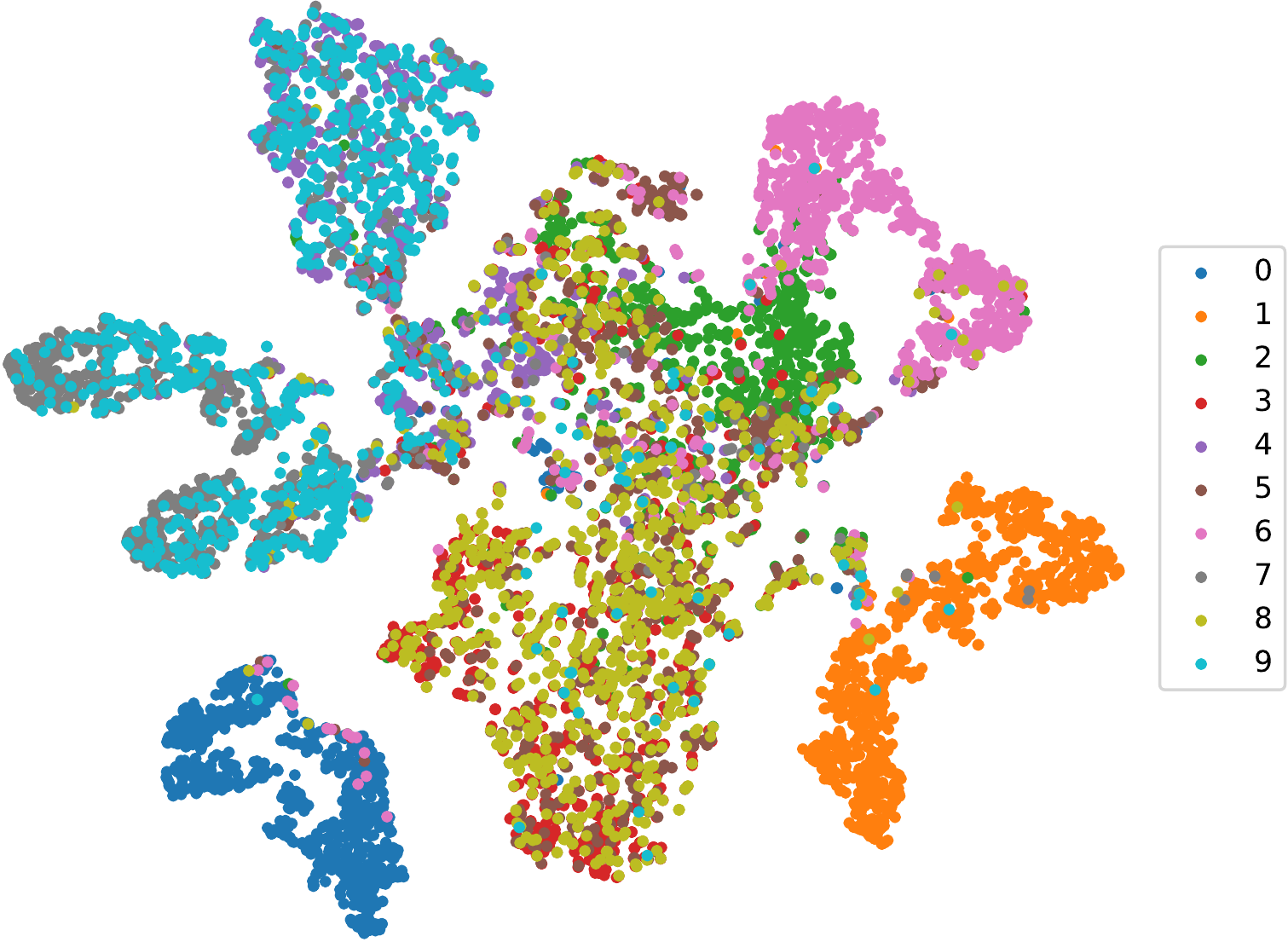} &
    \includegraphics[height=0.69in,trim={0 0 45px 0},clip]{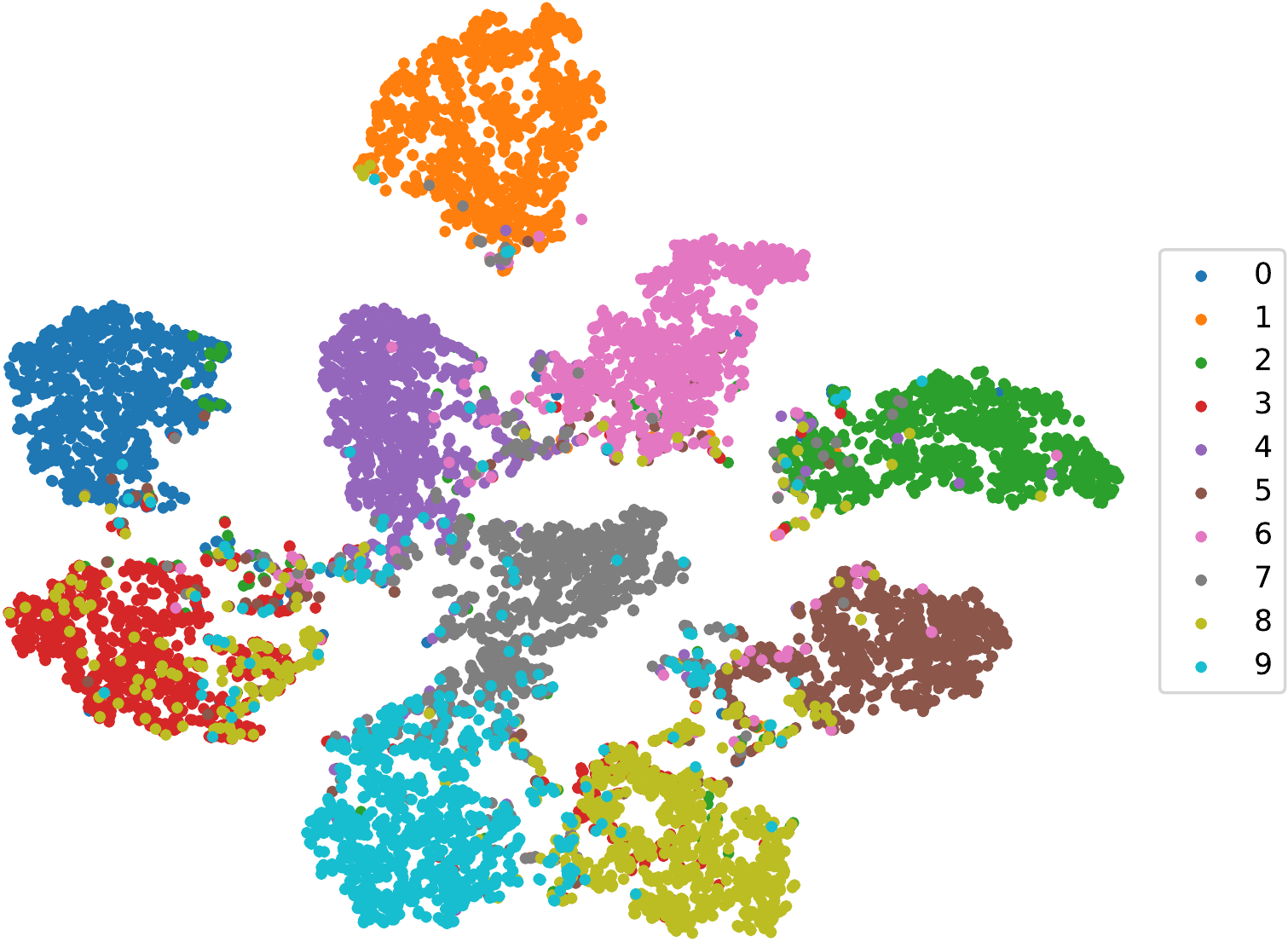} &
    \includegraphics[height=0.69in,trim={0 0 45px 0},clip]{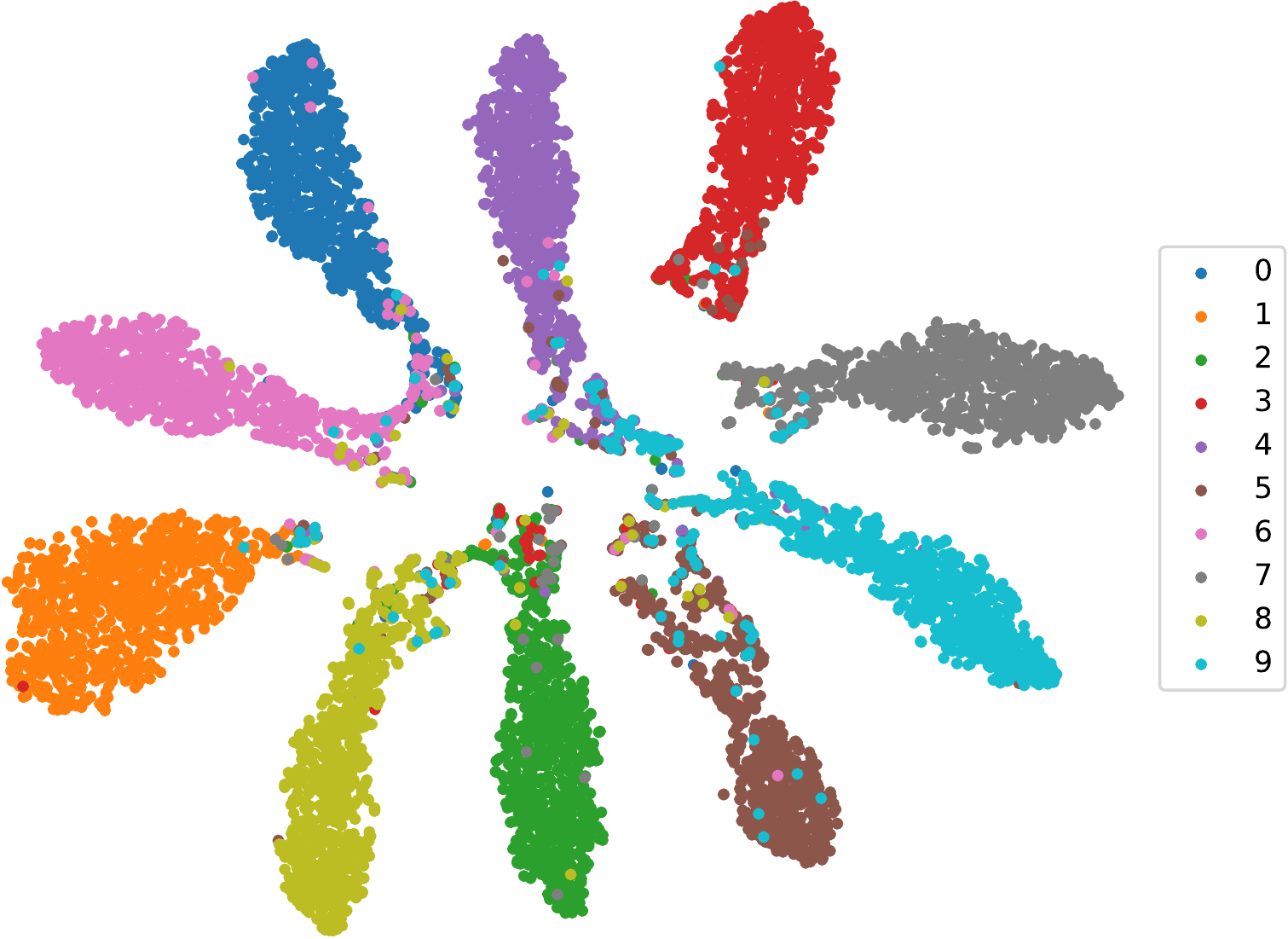} \\
  \end{tabular}
  \vskip0.5ex
  \caption{$t$-SNE plots for CCA methods on our variation of MNIST. Each method was used to compute projections for the two views (left and right sides of the images) using 10,000 training examples. The plots show a visualization of the projection for the left view with each digit colored differently. 
  \texttt{TOCCA-SD} and \texttt{TOCCA-ND} (not shown) produced similar results to \texttt{TOCCA-W}.}
  \label{fig_sne_mnist}
  \vspace{-0.1in}
\end{figure}

Across all training set sizes, our \texttt{TOCCA} variations consistently exhibited good performance, e.g., increasing classification accuracy from 78.3\% to 86.7\% for $n=1{,}000$ with \texttt{TOCCA-SD}.  Increases in accuracy over \texttt{TOCCA-ND} were small, indicating that the different decorrelation schemes have only a small effect on this data set; the task-driven component is the main reason for the success of our method.  In particular, the classification accuracy with $n=1{,}000$ did better than the unsupervised DCCA method on $n=10{,}000$. Further, \texttt{TOCCA} with $n=300$ did better than linear methods on $n=50{,}000$, clearly showing the benefits of the proposed formulation. We also examined the CCA projections qualitatively via a 2D $t$-SNE embedding \citep{VanDerMaaten2008}. Fig.~\ref{fig_sne_mnist} shows the CCA projection of the left view for each method. As expected, the task-driven variant produced more clearly separated classes.

\subsection{Regularization for cancer classification}

In this experiment, we address the following question: Given two views available for 
training but only one at test time, does the additional view help to regularize the model?

We study this question \mn{using 1,003 patient samples with} image and genomic data from CBCS\footnote{{\footnotesize \url{http://cbcs.web.unc.edu/for-researchers/}}} \citep{Troester2018}.  Images consisted of four cores per patient from a tissue microarray that was stained with hematoxylin and eosin.  Image features were extracted using a VGG16 backbone \citep{Simonyan2015}, pre-trained on ImageNet, by taking the mean of the 512D output of the fourth set of conv. layers across the tissue region and further averaging across all core images for the same patient. For gene expression (GE), we used the set of 50 genes in the PAM50 array \citep{Parker2009}. The data set was randomly split into half for training and one quarter for validation/testing; we report the mean over eight cross-validation runs. Classification tasks included predicting (1) Basal vs. non-Basal genomic subtype using images, which is typically done from GE, and (2) predicting grade 1 vs. 3 from GE, typically done from images.  \hdc{This is not a multi-task classification setup; it is a means for one view to stabilize the representation of the other.}

We tested different classifier training methods when only one view was available at test time: \mbox{a) a} linear SVM trained on one view, \mbox{b) a} \hdc{deep} NN trained on one view \hdc{using the same architecture as the lower layers of \texttt{TOCCA}}, \mbox{c) CCAL-$\mathcal{L}_{\text{ce}}$} trained on both views, \mbox{d) \texttt{TOCCA}} trained on both views. Table \ref{tbl_cbcs_image} lists the classification accuracy for each method and task. When predicting genomic subtype Basal from images, all our methods showed an improvement in classification accuracy; the best result was with \texttt{TOCCA-W}, which produced a 2.2\% improvement.  For predicting grade from GE, all our methods again improved the accuracy -- by up to 3.2\% with \texttt{TOCCA-W}.  These results show that having additional information during training can boost performance at test time.
\begin{table}
  \centering
  \caption{Classification accuracy for different methods of predicting Basal genomic subtype from images or grade from gene expression. Linear SVM and DNN were trained on a single view, while all other methods were trained with both views. By regularizing with the second view during training, all \texttt{TOCCA} variants improved classification accuracy. The standard error is in parentheses.}
  \vskip1ex
  \label{tbl_cbcs_image}
  \adjustbox{width=5.5in}
  {
    \setlength\tabcolsep{0.05in}
    \begin{tabular}{cc}
    \begin{tabular}{lcccc}
    \toprule
    \textbf{Method} & \textbf{Training data} & \textbf{Test data} & \textbf{Task} & \textbf{Accuracy} \\
    \midrule
    Linear SVM & Image only & Image & Basal & 0.777 (0.003) \\
    NN & Image only & Image & Basal & 0.808 (0.006) \\
    CCAL-$\mathcal{L}_{\text{ce}}$ & Image+GE & Image & Basal & 0.807 (0.008) \\
    \texttt{TOCCA-W} & Image+GE & Image & Basal & \bf 0.830 (0.006) \\
    \texttt{TOCCA-SD} & Image+GE & Image & Basal & 0.818 (0.006) \\
    \texttt{TOCCA-ND} & Image+GE & Image & Basal & 0.816 (0.004) \\
    \bottomrule
    \end{tabular}
    &
    \begin{tabular}{lcccc}
    \toprule
    \textbf{Method} & \textbf{Training data} & \textbf{Test data} & \textbf{Task} & \textbf{Accuracy} \\
    \midrule
    Linear SVM & GE only & GE & Grade & 0.832 (0.012) \\
    NN & GE only & GE & Grade & 0.830 (0.012) \\
    CCAL-$\mathcal{L}_{\text{ce}}$ & GE+image & GE & Grade & 0.804 (0.022) \\
    \texttt{TOCCA-W} & GE+image & GE & Grade & \bf 0.862 (0.013) \\
    \texttt{TOCCA-SD} & GE+image & GE & Grade & 0.856 (0.011) \\
    \texttt{TOCCA-ND} & GE+image & GE & Grade & 0.856 (0.011) \\
    \bottomrule
    \end{tabular}
    \end{tabular}
  }
  \vspace{-0.15in}
\end{table}
Notably, this experiment used a static set of pre-trained VGG16 image features in order to assess the utility of the method. The network itself could be fine-tuned end-to-end with our \texttt{TOCCA} model, providing an easy opportunity for data augmentation and likely further improvements in classification accuracy.

\subsection{Semi-supervised learning for speech recognition}

Our final experiments use speech data from XRMB, consisting of simultaneously recorded acoustic and articulatory measurements.  Prior work has shown that CCA-based algorithms can improve phonetic recognition \citep{Wang2015_ICASSP,Wang2015_ICML,Wang2016,Dorfer2016}.  The 45 speakers were split into 35 for training, 2 for validation, and 8 for testing -- a total of 1,429,236 samples for training, 85,297 for validation, and 111,314 for testing.\footnote{{\footnotesize \url{http://ttic.uchicago.edu/~klivescu/XRMB_data/full/README}}}  The acoustic features are 112D and the articulatory ones are 273D.  We removed the per-speaker mean \& variance for both views. Samples are annotated with one of 38 phonetic labels.

\renewcommand{\figurename}{Table}
\begin{wrapfigure}{r}{0.48\linewidth}
\vspace{-15pt}
\begin{small}
\caption{XRMB classification results.\label{tbl_xrmb}}
\centering
\begin{tabular}{lcc}
    \toprule
    \textbf{Method} & \textbf{Task} & \textbf{Accuracy} \\
    \midrule
    Baseline 	& - & 0.591 \\
    CCA 		& - & 0.589 \\
    RCCA 		& - & 0.588 \\
    DCCA 		& - & 0.620 \\
    SoftCCA 		& - & 0.635 \\
    Joint DCCA/DeepLDA & LDA & 0.633 \\
    CCAL-$\mathcal{L}_{\text{ce}}$ & Softmax & 0.642 \\
    \hline
    \texttt{TOCCA-W} & LDA & 0.710 \\
    \texttt{TOCCA-SD} & LDA & 0.677 \\
    \texttt{TOCCA-ND} & LDA & 0.677 \\
    \texttt{TOCCA-W} & Softmax & \bf 0.795 \\
    \texttt{TOCCA-SD} & Softmax & 0.785 \\
    \texttt{TOCCA-ND} & Softmax & 0.785 \\
    \bottomrule
    \end{tabular}
    \end{small}
    \vspace{-10pt}
\end{wrapfigure}
\renewcommand{\figurename}{Figure}

Our task on this data set was representation learning for multi-view prediction -- that is, using both views of data to learn a \hdc{shared} discriminative representation.  We trained each model using both views and their labels.  To test each CCA model, we followed prior work and concatenated the original input features from both views with the projections from both views.  Due to the large training set size, we used a Linear Discriminant Analysis (LDA) classifier for efficiency. The same construction was used at test time. This setup was used to assess whether a task-optimal DCCA model can improve discriminative power.  We tested \texttt{TOCCA} with a task-driven loss of LDA \citep{Dorfer2016_lda} or softmax to demonstrate the flexibility of our model.

\renewcommand{\figurename}{Table}
\begin{wrapfigure}{r}{0.7\linewidth}
\vspace{-25pt}
\begin{small}
\caption{Semi-supervised classification results on XRMB using \texttt{TOCCA-W}.\label{tbl_xrmb_semi}}
\begin{center}
\vspace{-0.1in}
\begin{tabular}{lc}
    \toprule
    \textbf{Labeled data} & \textbf{Accuracy} \\
    \midrule
    100\% & 0.795 \\
    30\% & 0.762 \\
    10\% & 0.745 \\
    3\% & 0.684 \\
    1\% & 0.637 \\
    \bottomrule
    \end{tabular}
    \end{center}
    \end{small}
    \vspace{-25pt}
\end{wrapfigure}
\renewcommand{\figurename}{Figure}
\hdc{We compared the discriminability of a variety of methods to learn a shared latent representation.} Table \ref{tbl_xrmb} lists the classification results with a 
baseline that used only the original input features for LDA. 
Although deep methods, i.e., DCCA and SoftCCA, improved upon the linear methods, all \texttt{TOCCA} variations significantly outperformed previous state-of-the-art techniques. Using softmax consistently beat LDA by a large margin.  \texttt{TOCCA-SD} and \texttt{TOCCA-ND} produced equivalent results as a weight of $0$ on the decorrelation term performed best. However, \texttt{TOCCA-W} showed the best result with an improvement of 15\% over the best alternative method.

\texttt{TOCCA} can also be used in a \emph{semi-supervised} manner when labels are available for only some samples. Table \ref{tbl_xrmb_semi} lists the results for \texttt{TOCCA-W} in this setting.  With 0\% labeled data, the result would be similar to DCCA. Notably, a large improvement over the unsupervised results in Table \ref{tbl_xrmb} is seen even with labels for only 10\% of the training samples.

\section{Discussion}

We proposed a method to find a shared latent space that is also discriminative by adding a task-driven component to deep CCA while enabling end-to-end training.  This was accomplished by replacing the CCA projection with $\ell_2$ distance minimization and orthogonality constraints on the activations, and was implemented in three different ways. \texttt{TOCCA}-W or \texttt{TOCCA-SD} performed the best, dependent on the data set -- both of which include some means of decorrelation to provide an extra regularizing effect to the model and thereby outperforming \texttt{TOCCA-ND}.  

\texttt{TOCCA} showed large improvements over state-of-the-art in cross-view classification accuracy on MNIST and significantly increased robustness when the training set size was small.  On CBCS, \texttt{TOCCA} provided a regularizing effect when both views were available for training but only one at test time.  \texttt{TOCCA} also produced a large increase over state-of-the-art for multi-view representation learning on a much larger data set, XRMB.  On this data set we also demonstrated a semi-supervised approach to get a large increase in classification accuracy with only a small proportion of the labels.  Using a similar technique, our method could be applied when some samples are missing a second view.

Classification tasks using a softmax operation or LDA were explored in this work; however, the formulation presented can also be used with other tasks such as regression or clustering.  Another possible avenue for future work entails extracting components shared by both views as well as individual components.  This approach has been developed for dictionary learning \citep{Lock2013,Ray2014_bioinformatics,Feng2018} but could be extended to deep CCA-based methods.  Finally, we have yet to apply data augmentation to the proposed framework; this could provide a significant benefit for small training sets.


\small

\bibliography{references}
\bibliographystyle{unsrt}

\beginsupplement

\section*{Supplementary Material}

This supplementary material includes additional details on our \texttt{TOCCA} algorithm and experiments, including 1) a comparison of our formulation with other related CCA approaches, 2) pseudocode for the ZCA whitening algorithm used by \texttt{TOCCA-W}, 3) details on hyperparameter selection, and 4) training runtime experiments.

\subsection*{Comparison of \texttt{TOCCA} with related algorithms}

Table \ref{tbl_methods} compares our three \texttt{TOCCA} formulations with other related linear and deep CCA methods.

\begin{table}[H]
  \centering
  \caption{A comparison of our proposed task-optimal deep CCA methods with other related ones from the literature: DCCA \citep{Andrew2013}, SoftCCA \citep{Chang2018}, CCAL-$\mathcal{L}_{\text{rank}}$ \citep{Dorfer2018}.  CCAL-$\mathcal{L}_{\text{rank}}$ uses a pairwise ranking loss with cosine similarity to identify matching and non-matching samples for image retrieval -- not classification.  $A_1$ and $A_2$ are mean centered outputs from two feed-forward networks.  $\Sigma = A^T A$ is computed from a single (large) batch (used in DCCA); $\hat{\Sigma}$ is computed as a running mean over batches (for all other methods).  $f_{\text{task}}(A;\theta_{\text{task}})$ is a task-specific function with parameters $\theta_{\text{task}}$, e.g., a softmax operation for classification.}
  \label{tbl_methods}
  \vskip0.5ex
  {
    \scalebox{0.64}{
    \setlength\tabcolsep{0.02in}
    \begin{tabular}{lrll}
    \toprule
    {\bf Method} & \multicolumn{2}{l}{{\bf Objective}} \\
    \midrule
    CCA & \multicolumn{2}{c}{$-\text{tr}(W_1^T \Sigma_{12} W_2)$} & s.t.  $W_1^T \Sigma_1 W_1 = W_2^T \Sigma_2 W^2 = I$ \\
    DCCA & \multicolumn{2}{c}{$-||\Sigma_1^{-1/2} \Sigma_{12} \Sigma_2^{-1/2}||_{\text{tr}}$} & where $||T||_{\text{tr}} = \text{tr}(T^T T)^{1/2}$  (TNO, equivalent to CCA objective) \\
    & & & $\text{CCA}(W_1^T A_1, W_2^T A_2)$ computed after optimization complete \\
    SoftCCA &   & ~~~~$\mathunderline{Cerulean}{\mathcal{L}_{\ell_2 \text{ dist} }(A_1,A_2)} + \lambda ~~\left( \mathunderline{Orchid}{\mathcal{L}_{\text{Decorr}}(A_1)} + \mathunderline{Orchid}{\mathcal{L}_{\text{Decorr}}(A_2)} \right)$ \\
    CCAL-$\mathcal{L}_{\text{rank}}$ & \multicolumn{1}{l}{$\mathcal{L}_{\text{rank}}(B_1,B_2)$} & & where $B_1,B_2 = \text{CCA}(A_1,A_2)$, $\mathcal{L}_{\text{rank}}$ is pairwise ranking loss \\
    \texttt{TOCCA-W} & $\mathunderline{SeaGreen}{\text{Task}(B_1,B_2,Y)} +$ &$\lambda~~ \mathunderline{Cerulean}{\mathcal{L}_{\ell_2 \text{ dist} }(B_1,B_2)}$ & where $\mathunderline{Maroon}{B_1 = U_1 A_1, B_2 = U_2 A_2 \text{ s.t. } B_1^T B_1 = B_2^T B_2 = I}$ \\
    \texttt{TOCCA-SD} & $\mathunderline{SeaGreen}{\text{Task}(A_1,A_2,Y)} +$ &$\lambda_1 \mathunderline{Cerulean}{\mathcal{L}_{\ell_2 \text{ dist} }(A_1,A_2)} + \lambda_2 \left( \mathunderline{Orchid}{\mathcal{L}_{\text{Decorr}}(A_1)} + \mathunderline{Orchid}{\mathcal{L}_{\text{Decorr}}(A_2)} \right)$ & \multicolumn{1}{c}{\textcolor{Maroon}{Whitening}} \\
    \texttt{TOCCA-ND}~~ & $\mathunderline{SeaGreen}{\text{Task}(A_1,A_2,Y)} +$ &$\lambda~~ \mathunderline{Cerulean}{\mathcal{L}_{\ell_2 \text{ dist} }(A_1,A_2)}$ \\
    \midrule
    \multicolumn{4}{l}{\text{\bf Loss functions}} \\
    \midrule
    $\ell_2$ dist & \multicolumn{3}{l}{$\mathunderline{Cerulean}{\mathcal{L}_{\ell_2 \text{ dist}}(A_1,A_2)} = ||A_1 - A_2||^2_F$} \\
    Decorr & \multicolumn{2}{l}{$\mathunderline{Orchid}{\mathcal{L}_{\text{Decorr}}(A)} = \sum_{i \neq j} |\hat{\Sigma}_{i,j}|$} & where $\hat{\Sigma}$ is running mean across batches of $\Sigma = A^T A$  \\
    Task & \multicolumn{2}{l}{$\mathunderline{SeaGreen}{\text{Task}(A_1,A_2,Y)} = \mathcal{L}_{\text{task}}(f_{\text{task}}(A_1;\theta_{\text{task}}),Y) + \mathcal{L}_{\text{task}}(f_{\text{task}}(A_2;\theta_{\text{task}}),Y)$} & where $\mathcal{L}_{task}$ can be cross-entropy or any other task-driven loss \\
    \bottomrule
  \end{tabular}
    }
    }
\end{table}


\subsection*{Algorithm for whitening}

Pseudocode for ZCA whitening used to achieve orthogonality in our \texttt{TOCCA-W} implementation is shown in Algorithm \ref{algo_zca}.

\begin{algorithm}[H]
\caption{Whitening layer for orthogonality.}
\label{algo_zca}
\begin{algorithmic}
  \STATE {\bfseries Input:} activations $A \epsilon \mathbb{R}^{d_o \times n}$\\
  \STATE {\bfseries Hyperparameters:} batch size $m$, momentum $\alpha$
  \STATE {\bfseries Parameters of layer:} mean $\mu$, covariance $\Sigma$
  \IF{training}
  \STATE $\mu \gets \alpha \mu + (1-\alpha) \frac{1}{m} A ~1_{n \times 1}$   \COMMENT{Update mean}
  \STATE $\bar{A} = A - \mu$ \COMMENT{Mean center data}
  \STATE $\Sigma \gets \alpha \Sigma + (1-\alpha) \frac{1}{m-1} \bar{A}_1 \bar{A}_2^T$ \COMMENT{Update covariance}
  \STATE $\hat{\Sigma} \gets \Sigma + \epsilon I$ \COMMENT{Add $\epsilon I$ for numerical stability}
  \STATE $\Lambda,V \gets \text{eig}(\hat{\Sigma})$ \COMMENT{Compute eigendecomposition}
  \STATE $U \gets V \Lambda^{-1/2} V^T$ \COMMENT{Compute transformation matrix}
  \ELSE
  \STATE $\bar{A} \gets A - \mu$ \COMMENT{Mean center data}
  \ENDIF
  \STATE $B \gets U \bar{A}$ \COMMENT{Apply ZCA whitening transform}
  \STATE \textbf{return}  $B$
\end{algorithmic}
\end{algorithm}


\subsection*{Implementation details: hyperparameters}

A random search over hyperparameters was used to train our methods.  The hyperparameter settings and ranges for each data set are provided in Table \ref{tbl_hyperparameters}.

\begin{table}[H]
  \centering
  \caption{Hyperparameter settings and search ranges for the experiments on each data set.}
  \label{tbl_hyperparameters}
  \begin{tabular}{lccc}
    \toprule
    \textbf{Hyperparameter} & \textbf{MNIST} & \textbf{CBCS} & \textbf{XRMB} \\
    \midrule
    Hidden layers & 4 & [0,4] & 4 \\
    Hidden layer size & 500 & 200 & 1,000 \\
    Output layer size & 50 & 50 & 112 \\
    Loss function weight $\lambda$ & $[10^0,10^{-4}]$ & $[10^1,10^{-5}]$ & $[10^1,10^{-5}]$ \\
    Momentum $\alpha$ & 0.99 & 0.99 & 0.99 \\
    $\ell_2$ regularizer & $[10^{-3},10^{-6}]$, 0 & $[10^{-2},10^{-5}]$, 0  & $[10^{-3},10^{-7}]$, 0 \\
    Soft decorrelation regularizer & $[10^0,10^{-5}]$ & $[10^0,10^{-5}]$ & $[10^0,10^{-5}]$ \\
    Batch size & 32 & 100 & 50,000 \\
    Learning rate & $[10^{-2},10^{-4}]$ & $[10^{-1},10^{-3}]$ & $[10^{0},10^{-4}]$ \\
    Epochs & 200 & 400 & 100 \\
    \bottomrule
  \end{tabular}
\end{table}

\subsection*{Runtime experiments}

The computational complexity of \texttt{TOCCA-W} is greater than that of \texttt{TOCCA-SD} due to the eigendecomposition operation (see \S\ref{sec_methods} in the main article); however, this extra computation is only carried out once per batch.  A runtime comparison of the two methods on all three data sets is provided in Table \ref{tbl_runtime}.

\begin{table}[H]
  \centering
  \caption{Training runtime for each data set.}
  \label{tbl_runtime}
  \begin{tabular}{lcccc}
    \toprule
    \textbf{Data set} & \textbf{Batch size} & \textbf{Epochs} & \textbf{\texttt{TOCCA-W}} & \textbf{\texttt{TOCCA-SD}} \\
    \midrule
    MNIST & 100 & 200 & 488 s & 418 s \\
    MNIST & 30 & 200 & 1071 s & 1036 s \\
    CBCS & 100 & 400 & 103 s & 104 s \\
    XRMB & 50,000 & 100 & 3056 s & 3446 s \\
    \bottomrule
  \end{tabular}
\end{table}

\end{document}